# PROFICIENCY COMPARISON OFLADTREE AND REPTREE CLASSIFIERS FOR CREDIT RISK FORECAST

Lakshmi Devasena C

Dept. of Operations & Systems, ISB Hyderabad, IFHE University

*Abstract*

Predicting the Credit Defaulter is a perilous task of Financial Industries like Banks. Ascertainingnonpayer before giving loan is a significant and conflict-ridden task of the Banker. Classification techniques are the better choice for predictive analysis like finding the claimant, whether he/she is an unpretentious customer or a cheat. Defining the outstanding classifier is a risky assignment for any industrialist like a banker. This allow computer science researchers to drill down efficient research works through evaluating different classifiers and finding out the best classifier for such predictive problems. This research work investigates the productivity of LADTree Classifier and REPTree Classifier for the credit risk prediction and compares their fitness through various measures. German credit dataset has been taken and used to predict the credit risk with a help of open source machine learning tool.

## 1.INTRODUCTION

The enormous volume of transactions made information processing automation an invigorating factor for high quality standards, cost reduction, with high speed results. Data analysis automation and result of the relevant successes produced by state-of-the art computer algorithms have changed the opinions of many misanthropists. In the past, people thought that financial market analysis necessitates intuition, knowledge and experience and speculated how this job could be automated. Conversely, growth of scientific and technological advances, achieved the automation of financial market analysis. In recent days, credit defaulter prediction and credit risk evaluation have fascinated great deal of interests from regulators, practitioners, and theorists, in the financial industry. Since, the credit score of an applicant could be calculated from the past giant database and the demographic data, it needs automation. Automation of credit risk forecastcan be achieved using classification techniques. Selecting the classifier, which envisages credit risk in an efficient manner, is an imperative and critical task. This research work appraises the credit risk performance of two diverse classifiers, namely, REP Tree Classifier and LAD Tree Classifier and compares their accuracyofcredit risk prediction.

International Journal on Computational Sciences & Applications (IJCSA) Vol.5, No.1, February 2015

## 2. LITERATURE REVIEW

There are many research works made to predict credit risk using wide-ranging computing techniques. In [1], a neural network based algorithm for automatic provisioning to credit risk scrutiny in a real world problem is presented. An assimilated back propagation neural network (BPNN) with the customary discriminant analysis approach used to discover the performance of credit scoring is given in [2]. A comparative study of corporate credit rating analysis using back propagation neural network (BPNN) and support vector machines (SVM) is described in [3]. An uncorrelated maximization algorithm within a triple-phase neural network ensemble technique for credit risk evaluation to differentiate good creditors from bad ones are elucidated in [4]. An application of artificial neural network to credit risk assessment using two altered architectures are deliberated in [5]. Credit risk investigation using diverse Data Mining models like C4.5, NN, BP, RIPPER, LR and SMO islikened in [6]. The credit risk of a Tunisian bank through modeling the non-payment risk of its commercial loans is analyzed in [7]. Credit risk valuation using six stage neural network ensemble learning approach is argued in [8]. A modeling framework for credit calculation models is erected using different modeling procedures is explained and its performance is analyzed in [9]. Hybrid method for assessing credit risk using Kolmogorove-Smirnov test, Fuzzy Expert system and DEMATEL method is enlightened in [10]. An Artificial Neural Network centeredmethodology for Credit Risk supervision is proposed in [11]. Artificial neural networks using Feed-forward back propagation neural network and business rules to correctly determine credit defaulter is proposed in [12]. The performance comparison of Memory based classifiers for credit risk investigation is experimented and précised in [13]. The performance comparison between Instance Based and K Star Classifiers for Credit Risk Inspection is accomplished and pronounced in [14]. The performance comparison among Sequential Minimal Optimization and Logistic Classifiers for Credit Risk Calculation is specified in [15]. The performance comparisonbetween Multilayer Perceptron and SMO Classifier for Credit Risk appraisal is described in [16]. The performance comparison between JRip and PART Classifier for Credit Risk Estimation is explored in [17]. Tree based classifiers are easier to interpret and explain. That's the reason; this research work randomly taken LAD Tree classifier and REP Tree Classifier which are used in various optimization literatures [23] for efficiency comparison. LAD Tree and REP Tree Classifiers have already used in various domain of problem like Forensic Mining [24], Intrusion Detection system [25] and [36], Non-spatial Data Classification [26], Classification for Bank Direct Marketing [27], Global land cover classification [28], Medical Data Classification [29], [30] and [34] , Automatic classification of active objects from large categories [31], Micro array dataset classification and analysis [32], Classification of Strom type from weather radar reflectivity [33], Identification of Link Spam in Web search Engines [35], etc.

## 3. DATASET USED

The German credit data [18] is used to evaluate the performance of Logistic classifier and Partial Decision Tree Classifier for credit risk prediction. This data setcontains 20 attributes, namely, Duration, Credit History, Checking Status, Purpose, Credit Amount, Employment, Installment Commitment, Saving Status, Personal Status, Other parties, Property magnitude, Age, resident since, Other payment plans, existing credits, job, Housing, No. of dependents, Foreign worker and Own Phone. The data set comprises 1000 instances of client credit data with class detail. It discriminates the records into two classes, namely, good and bad.

## 4. METHODOLOGY USED

In this research work, two diversetree based classifiers namely, LAD Tree Classifierand REP Tree Classifier are compared for proficiencyassessment of credit risk estimation.

### 4.1.LAD Tree Classifier

LAD Tree builds a classifier for binary target variable based on learning a logical expression that can discriminate between positive and negative samples in a data set. The construction of LAD model for a given data set typically involves the generation of large set patterns and the selection of a subset of them that satisfies the above assumption that a binary point covered by some positive patterns, but not covered by any negative pattern is positive, and similarly, a binary point covered by some negative patterns, but not covered by positive pattern is negative, such that each pattern in the model satisfies certain requirements in terms of prevalence and homogeneity[23]. LAD Tree Classifier generates a multi-class alternating decision tree using the Logit Boost strategy. The LAD Tree algorithm applies logistic boosting algorithm in order to induce an alternating decision tree. In this algorithm, a single attribute test is chosen as a splitter node for the tree at each iteration. For each training instance, working response and weights are calculated and stored on a per-class basis. Then, it fits the working response to the mean value of the instances, in a particular subset, by minimizing the least-squares value between them. IN this algorithm, trees for the different classes are grown in parallel. Once all the trees have been constructed, then it merges the trees into a final model. Advantage of this classifier is the size of the tree cannot outgrow the combined size of the individual trees [19].

### 4.2.REP Tree Classifier

Reduces Error Pruning (REP) Tree Classifier is a fast decision tree learning algorithm and is based on the principle of computing the information gain with entropy and minimizing the error arising from variance [20]. This algorithm is first recommended in [21]. REP Tree applies regression tree logic and generates multiple trees in altered iterations. Afterwards it picks best one from all spawned trees. This algorithm constructs the regression/decision tree using variance and information gain. Also, this algorithm prunes the tree using reduced-error pruning with back fitting method. At the beginning of the model preparation, it sorts the values of numeric attributes once. As in C4.5 Algorithm, this algorithm also deals the missing values by splitting the corresponding instances into pieces.[22].

## 5.PERFORMANCE MEASURES USED

Variousscales are used to gauge the performance of the classifiers.

### Classification Accuracy

Any classifier could have an error rate and it may fail to categorize correctly. Classification accuracy is calculated as Correctly classified instances divided by Total number of instances multiplied by 100.

### Mean Absolute Error

International Journal on Computational Sciences & Applications (IJCSA) Vol.5, No.1, February 2015

Mean absolute error is the average of the variance between predicted and actual value in all test cases. It is a good measure to gauge the performance.

### Root Mean Square Error

Root mean squared error is used to scaledissimilarities between values actually perceived and the values predicted by the model. It is determined by taking the square root of the mean square error.

### Confusion Matrix

A confusion matrix encompasses information about actual and predicted groupings done by a classification system.

## 6. RESULTS AND DISCUSSION

Open source machine learning tool is used to experiment the performance of LAD Tree and REP Tree Classifiers. The performance is tested out using the Training set as well as using different Cross Validation methods. The class is arrived by considering all 20 attributes of the dataset.

### 6.1. Performance of LAD Tree Classifier

The overall assessment summary of LAD Tree Classifier using training set and different cross validation methods is given in Table I. The performance of LAD Tree Classifier in terms of Correctly Classified Instances and Classification Accuracy is shown in Fig. 1and Fig. 2. The confusion matrix for different test mode is given in Table II to Table VI. LAD Tree Classifier gives 76.1% accuracy for the training data set. Various cross validation methods are used to check its actual performance. On an average, it gives around 70.7% of accuracy for credit risk estimation.

TABLE I
LAD TREE CLASSIFIER OVERALL EVALUATION SUMMARY

| Test Mode | Correctly Classified Instances | Incorrectly Classified Instances | Accuracy | Mean Absolute Error | Root Mean Squared Error | Time Taken to Build Model (Sec) |
|---|---|---|---|---|---|---|
| **Training Set** | 761 | 239 | 76.1% | 0.3236 | 0.3953 | 2.64 |
| **5 Fold CV** | 702 | 298 | 70.2% | 0.3547 | 0.437 | 2.07 |
| **10 Fold CV** | 708 | 292 | 70.8% | 0.3494 | 0.4326 | 1.92 |
| **15 Fold CV** | 715 | 285 | 71.5% | 0.3545 | 0.4351 | 2.67 |
| **20 Fold CV** | 704 | 296 | 70.4% | 0.3559 | 0.437 | 1.43 |

TABLE II
CONFUSION MATRIX –LAD TREE CLASSIFIER (ON TRAINING DATASET)

| | Good | Bad | Actual (Total) |
|---|---|---|---|
| **Good** | 655 | 45 | 700 |
| **Bad** | 194 | 106 | 300 |
| **Predicted (Total)** | 849 | 151 | 1000 |

TABLE III
CONFUSION MATRIX – LAD TREE CLASSIFIER (5 FOLD CROSS VALIDATION)

| | Good | Bad | Actual (Total) |
|---|---|---|---|
| **Good** | 585 | 115 | 700 |
| **Bad** | 183 | 117 | 300 |
| **Predicted (Total)** | 768 | 232 | 1000 |

TABLE IV
CONFUSION MATRIX – LAD TREE CLASSIFIER (10 FOLD CROSS VALIDATION)

| | Good | Bad | Actual (Total) |
|---|---|---|---|
| **Good** | 597 | 103 | 700 |
| **Bad** | 189 | 111 | 300 |
| **Predicted (Total)** | 786 | 214 | 1000 |

TABLE V
CONFUSION MATRIX – LAD TREE CLASSIFIER (15 FOLD CROSS VALIDATION)

| | Good | Bad | Actual (Total) |
|---|---|---|---|
| **Good** | 614 | 86 | 700 |
| **Bad** | 199 | 101 | 300 |
| **Predicted (Total)** | 813 | 187 | 1000 |

TABLE VI
CONFUSION MATRIX –LAD TREE CLASSIFIER (20 FOLD CROSS VALIDATION)

| | Good | Bad | Actual (Total) |
|---|---|---|---|
| **Good** | 594 | 106 | 700 |
| **Bad** | 190 | 110 | 300 |
| **Predicted (Total)** | 784 | 216 | 1000 |

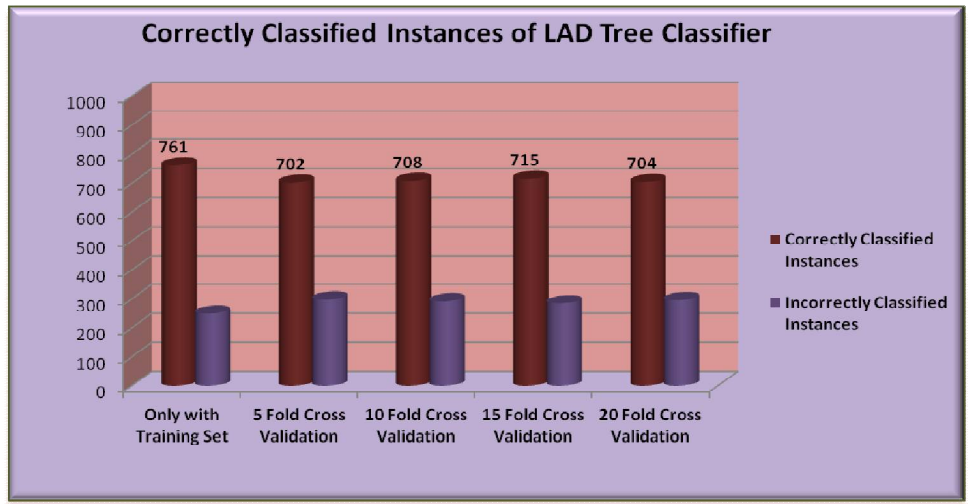

Fig. 1 Correctly Classified instances of LAD Tree Classifier

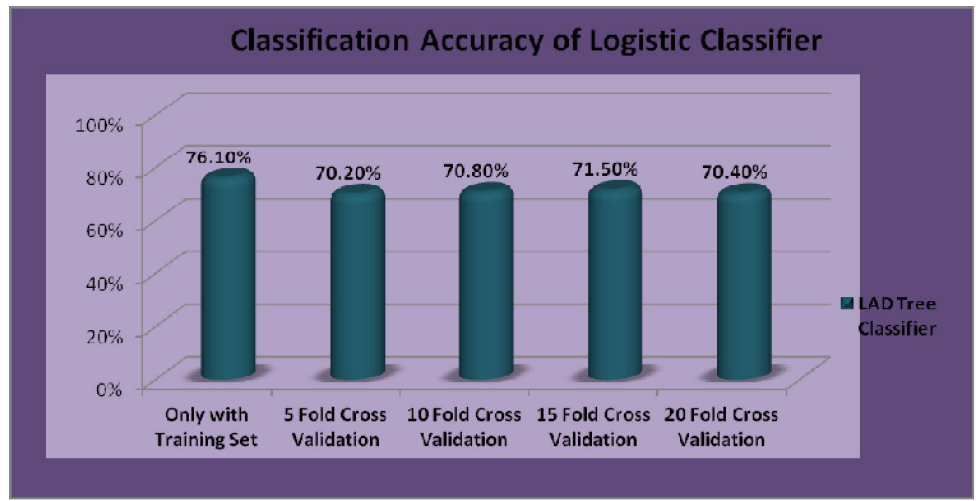

Fig. 2 Classification Accuracy of LAD Tree Classifier

## 6.2. Performance of REP Tree Classifier

The overall assessment summary of REP Tree Classifier using training set and different cross validation methods is given in Table VII. The performance of REP Tree Classifier in terms of Correctly Classified Instances and Classification Accuracy is shown in Fig. 3and Fig. 4. The confusion matrix for different test mode is given in Table VIII to Table XII. REP Tree Classifier gives 80% accuracy for the training data set. Various cross validation methods are used to check its actual performance. On an average, it gives around 71.9% of accuracy for credit risk estimation.

TABLE VIII
REP TREE CLASSIFIER COMPLETE EVALUATION SUMMARY

| Test Mode | Correctly Classified Instances | Incorrectly Classified Instances | Accuracy | Mean absolute error | Root Mean Squared Error | Time Taken to Build Model (Sec) |
|---|---|---|---|---|---|---|
| Training Set | 800 | 200 | 80% | 0.2905 | 0.3811 | 0.32 |
| 5 Fold CV | 717 | 283 | 71.7% | 0.3458 | 0.4437 | 0.78 |
| 10 Fold CV | 718 | 282 | 71.8% | 0.3417 | 0.4424 | 1.33 |
| 15 Fold CV | 726 | 274 | 72.6% | 0.3422 | 0.4382 | 0.16 |
| 20 Fold CV | 719 | 281 | 71.9% | 0.3368 | 0.4364 | 0.11 |

TABLE VIII
CONFUSION MATRIX – REP TREE CLASSIFIER (ON TRAINING DATASET)

|  | Good | Bad | Actual (Total) |
|---|---|---|---|
| Good | 649 | 51 | 700 |
| Bad | 149 | 151 | 300 |
| Predicted (Total) | 798 | 202 | 1000 |

TABLE IX
CONFUSION MATRIX – REP TREE CLASSIFIER (5 FOLD CROSS VALIDATION)

|  | Good | Bad | Actual (Total) |
|---|---|---|---|
| Good | 616 | 84 | 700 |
| Bad | 199 | 101 | 300 |
| Predicted (Total) | 815 | 185 | 1000 |

TABLE X
CONFUSION MATRIX – REP TREE CLASSIFIER (10 FOLD CROSS VALIDATION)

|  | Good | Bad | Actual (Total) |
|---|---|---|---|
| Good | 601 | 99 | 700 |
| Bad | 183 | 117 | 300 |
| Predicted (Total) | 784 | 216 | 1000 |

TABLE XI
CONFUSION MATRIX – REP TREE CLASSIFIER (15 FOLD CROSS VALIDATION)

|  | Good | Bad | Actual (Total) |
|---|---|---|---|
| Good | 612 | 88 | 700 |
| Bad | 186 | 114 | 300 |
| Predicted (Total) | 798 | 202 | 1000 |

TABLE XII
CONFUSION MATRIX – REP TREE CLASSIFIER (20 FOLD CROSS VALIDATION)

|  | Good | Bad | Actual (Total) |
|---|---|---|---|
| **Good** | 605 | 95 | 700 |
| **Bad** | 151 | 149 | 300 |
| **Predicted (Total)** | 756 | 244 | 1000 |

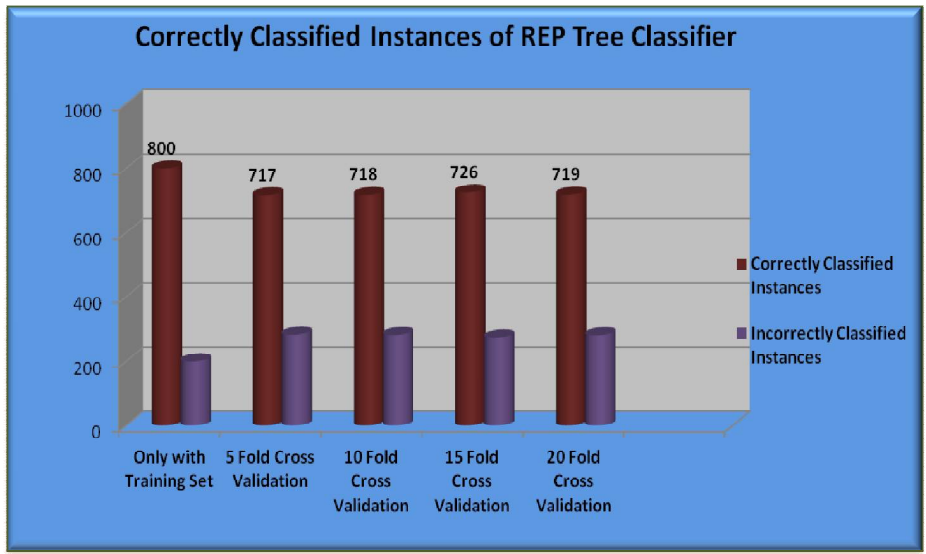

Fig. 3 Correctly Classified instances of REP Tree Classifier

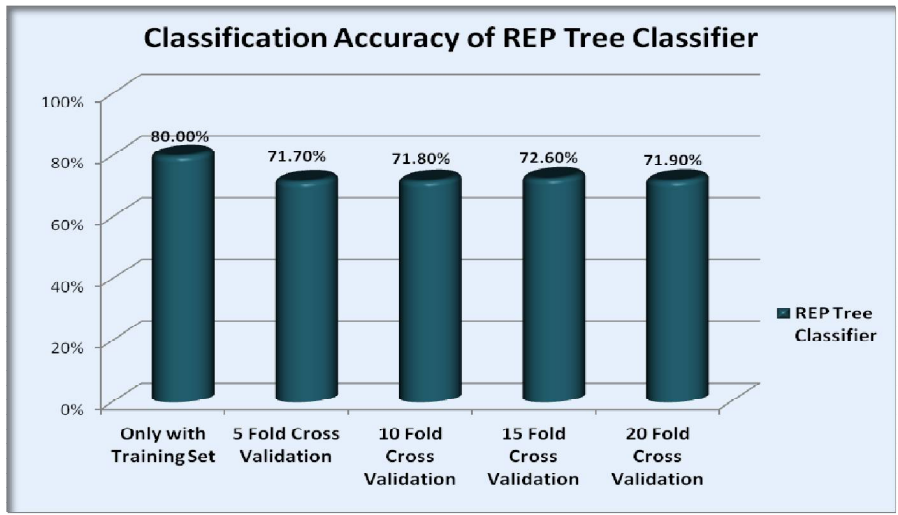

Fig. 4 Classification Accuracy of REP Tree Classifier

International Journal on Computational Sciences & Applications (IJCSA) Vol.5, No.1, February 2015

## 6.3 Comparison of LAD Tree Classifier and REP Tree Classifier

The comparison of performance between LAD Tree Classifier and REP Tree Classifier is depicted in Fig 5, and Fig. 6 in terms of Correctly Classified Instances and Classification Accuracy. The complete ranking is prepared based on correctly classified instances, classification accuracy, MAE and RMSE values and other statistics found using Training Set result and Cross Validation Techniques. Consequently, it is perceived that REP Tree classifier performs better than LAD Tree Classifier.

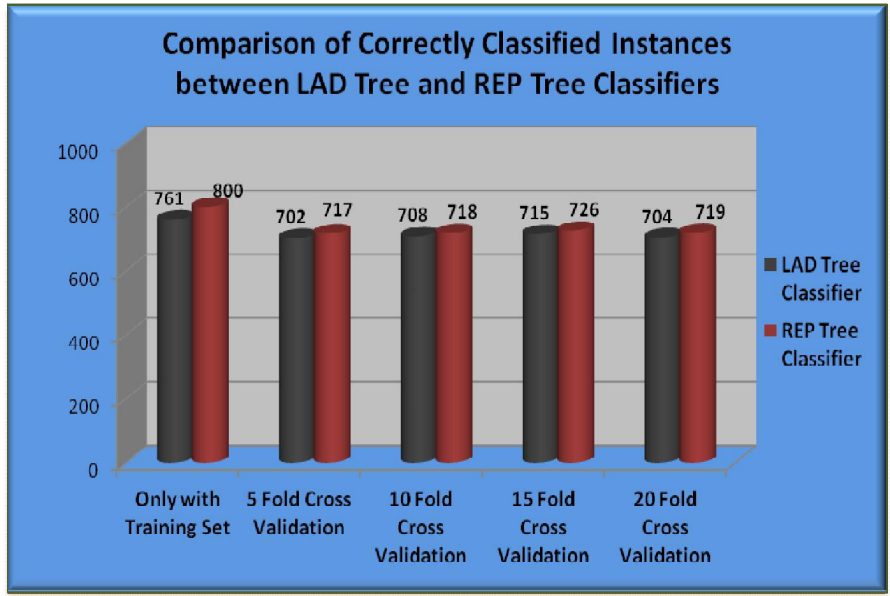

Fig. 5 Correctly Classified Instances Comparison between LAD Tree Classifier and REP Tree Classifier

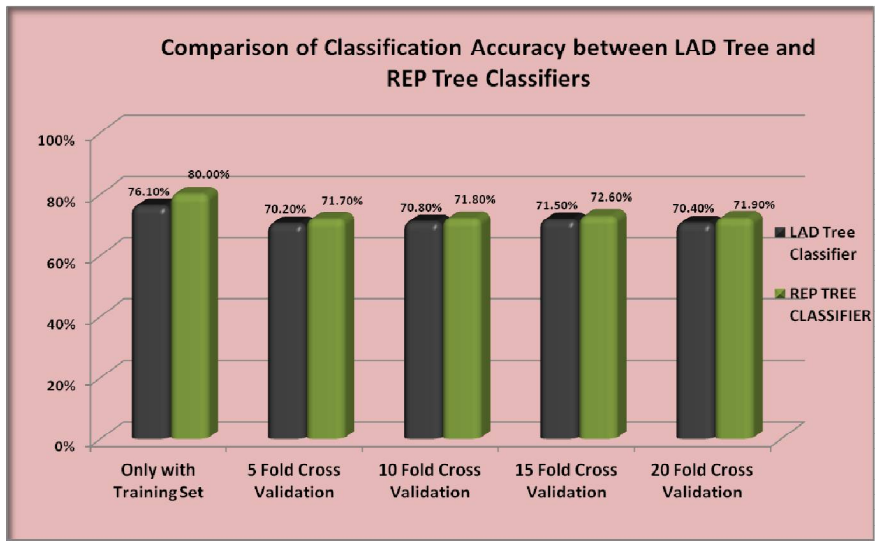

Fig. 6 Classification Accuracy Comparison between LAD Tree Classifier and REP Tree Classifier

## 7. CONCLUSION

This work investigated the efficiency of two different classifiers namely, LAD Tree Classifier andREP Tree Classifier for credit risk prediction. Testing is accomplished using the open source machine learning tool. Also, effectiveness comparison of both the classifiers has been done in view of different scales of performance evaluation. At last, it is observed that REP Tree Classifier performs better than LAD Tree Classifier for credit risk prediction by taking various measures including Classification accuracy and Time taken to build the model.

## ACKNOWLEDGMENT

The author expresses her deep gratitude to the Management of IBS Hyderabad, IFHE University and Operations & IT Department of IBS Hyderabad for constant support and motivation.